\documentclass[10pt,twocolumn,letterpaper]{article}

\usepackage{cvpr}              

\usepackage[dvipsnames]{xcolor}

\definecolor{cvprblue}{rgb}{0.21,0.49,0.74}
\usepackage[pagebackref,breaklinks,colorlinks,citecolor=cvprblue]{hyperref}
\usepackage{graphicx}
\usepackage{amsmath}
\usepackage{amssymb}
\usepackage{booktabs}
\usepackage{color}
\usepackage{multicol}
\usepackage{multirow}
\usepackage{array}

\author{
Ruiqi Wu\textsuperscript{1 2 *}\quad Liangyu Chen\textsuperscript{2}\quad Tong Yang\textsuperscript{2} \quad Chunle Guo\textsuperscript{1 $\dag$} \quad Chongyi Li\textsuperscript{1}\quad Xiangyu Zhang\textsuperscript{2}\\
\textsuperscript{1}VCIP, CS, Nankai University\quad
\textsuperscript{2}MEGVII Technology\\
{\tt\small wuruiqi@mail.nankai.edu.cn, }{\tt\small\{chenliangyu, yangtong, zhangxiangyu\}@megvii.com}\\
{\tt\small \{guochunle, lichongyi\}@nankai.edu.cn}\\
}

\newcommand{\methodname}{LAMP}
\begin{document}
\title{LAMP:  \underline{L}earn \underline{A} \underline{M}otion \underline{P}attern for Few-Shot-Based Video Generation}

\twocolumn[{
\renewcommand\twocolumn[1][]{#1}
\maketitle
\begin{center}
    \captionsetup{type=figure}
    \vspace{-0.8cm}
    \includegraphics[width=\textwidth]{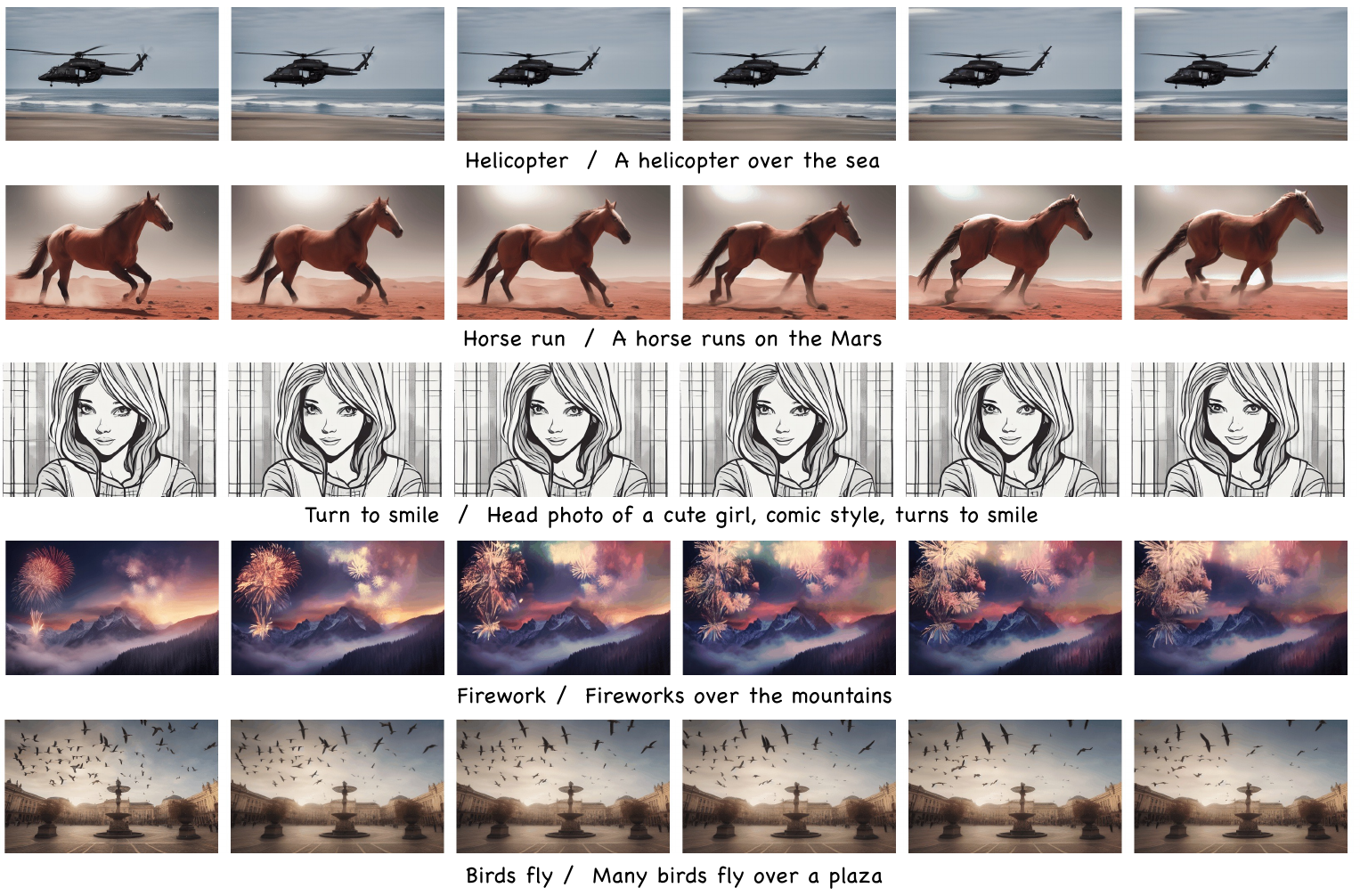}
    \vspace{-0.7cm}
    \captionof{figure}{\textbf{Our text-to-video results}. The motion prompts and video prompts are listed, respectively. Our \methodname~works effectively on diverse motions. The generated videos are temporal consistent and close to the video prompts. Moreover, two advantages of \methodname~can be reflected in the above results. (1) The proposed first-frame-conditioned training strategy allows us to use powerful SD-XL for first-frame generation, which is beneficial to producing highly detailed following frames. (2) Good semantic generalization properties of the diffusion model are preserved (\eg imposing smile's motion on unseen comic style) since our tuning way.}
    \label{fig:results}
\end{center}

}]

\begin{abstract}
%
With the impressive progress in diffusion-based text-to-image generation\let\thefootnote\relax\footnotetext{* This work is done during Ruiqi Wu's internship at MEGVII Technology.},  extending such powerful generative ability to text-to-video raises enormous attention.
\let\thefootnote\relax\footnotetext{$^\dag$ Correspondence author.}Existing methods either require large-scale text-video pairs and a large number of training resources or learn motions that are precisely aligned with template videos.
It is non-trivial to balance a trade-off between the degree of generation freedom and the resource costs for video generation.
In our study, we present a few-shot-based tuning framework, \textbf{\methodname},
which enables text-to-image diffusion model \textbf{L}earn \textbf{A} specific \textbf{M}otion \textbf{P}attern with 8 $\sim$16 videos on a single GPU.
Specifically, we design a first-frame-conditioned pipeline that uses an off-the-shelf text-to-image model for content generation so that our tuned video diffusion model mainly focuses on motion learning.
%
The well-developed text-to-image techniques can provide visually pleasing and diverse content as generation conditions, which highly improves video quality and generation freedom.
To capture the features of temporal dimension, we expand the pre-trained 2D convolution layers of the T2I model to our novel temporal-spatial motion learning layers and modify the attention blocks to the temporal level.
Additionally, we develop an effective inference trick, shared-noise sampling, which can improve the stability of videos with computational costs.
Our method can also be flexibly applied to other tasks, e.g. real-world image animation and video editing.
Extensive experiments demonstrate that \methodname~can effectively learn the motion pattern on limited data and generate high-quality videos.
The code and models are available at \href{https://rq-wu.github.io/projects/LAMP}{https://rq-wu.github.io/projects/LAMP}.
\end{abstract}    
\section{Introduction}
In recent years, generative models, particularly diffusion-based models~\cite{ho2020denoising, song2020denoising, song2020score}, have shown remarkable achievements in generating images from textual prompts, \ie text-to-image generation (T2I)~\cite{nichol2021glide, ramesh2022hierarchical, saharia2022photorealistic, rombach2022high, podell2023sdxl}.
Despite the success made in the T2I field, which has provided substantial technical groundwork, there remain large gaps in the development of text-to-video (T2V) generation: how to generate consistent frames and understand the motion patterns implicit in textural prompts.

Several recent works~\cite{ho2022imagen, singer2022make, zhou2022magicvideo, blattmann2023align, wang2023videocomposer, guo2023animatediff} try to bridge these gaps by directly training a diffusion-based T2V model using millions of text-video pairs.
These approaches facilitate a deeper understanding of the relationship between the video and the textural prompt.
However, the massive demand for labeled data and the heavy training burden are unaffordable for most researchers, constraining the development of this research line.
Another research line~\cite{molad2023dreamix, esser2023structure, wu2023tune, qi2023fatezero, yang2023rerender, geyer2023tokenflow} involves utilizing a video template and manipulating content using diffusion models while keeping the original motion. Those template-based methods are also known as video editing.
Although these methods can prove cost-effective, especially with the proposal of one-shot~\cite{wu2023tune} and even zero-shot~\cite{qi2023fatezero, yang2023rerender, geyer2023tokenflow} algorithms, the use of given video template significantly limits the generation freedom.
Besides, recent T2V-Zero~\cite{khachatryan2023text2video}  modifies the T2I diffusion models to generate consistent videos without training.
Nevertheless, it is challenging to transfer the text-image domain knowledge to the text-video domain in a zero-shot manner, resulting in the limitation of  T2V-Zero to generate similar-looking frames with random motions.

It is essential to achieve a trade-off between training burden and generation freedom while making models understand the motions.
Since the pretrained T2I diffusion model has good semantic comprehension guided by the prompts, it is reasonable that very little data is needed to make it understand the correspondence between prompts and motions and generate diverse videos.
Therefore, we attempt to explore a novel few-shot setting for the T2V task.
The new setting aims at tuning a T2I diffusion model to \textbf{L}earn \textbf{A} common \textbf{M}otion \textbf{P}attern from a small video set.

When tuning a T2I model to a T2V model in a few-shot manner, two issues need to be addressed.
(1) Due to the limited data amount, there is a risk of over-fitting the content within the video set. If the generated videos are similar to the video set, it undermines one of our core goals, namely generation freedom.
(2) The base operators of T2I diffusion models only work on spatial dimensions, which limits their ability to capture temporal information within videos.

With the two challenging issues, we propose a baseline method for few-shot T2V generation, named \textit{\textbf{\methodname}}.
Our solution to the first issue is the proposed \textbf{first-frame-conditioned pipeline}.
It decouples the T2V task into two sub-tasks,  generating the first frame by a pre-trained T2I model and predicting subsequent frames using our tuned video diffusion model.
The proposed pipeline seamlessly integrates the first frame as a condition without involving any additional model modification (\eg changing the data structure of inputs or adding new cross-attention layers).
Specifically, during training, we retain the first frame of the input video, adding noise and imposing the loss only on the subsequent frames.
Since the first frame provides the majority of the video's content, our model can focus on the relationship between the subsequent frames and the first frame, \ie the motion pattern rather than the contents.
%
%
During the inference, the first frame is generated by a pre-trained T2I model, such as SD-XL~\cite{podell2023sdxl}. 
We observe that a high quality of the first frame can boost the video generation performance through the proposed pipeline.
With the reference provided by the first frame, our model, which is based on Stable Diffusion v1.4 (SD v1.4)~\cite{rombach2022high}, can preserve the high-quality content generated by SD-XL throughout the video.
%
Facing the second issue, we design \textbf{temporal-spatial motion learning layers} to capture the features of temporal and spatial dimensions simultaneously.
%
Since predicting subsequent frames based on the first frame is required in the proposed pipeline, we modified the base operator based on the video prediction tasks~\cite{lotter2016deep, hu2023dynamic}, which will be introduced in Sec.~\ref{sec:architecture}.
As in previous works~\cite{wu2023tune, khachatryan2023text2video}, we modify the attention layers to build effective communication between frames.
Moreover, we adopt a \textbf{shared-noise sampling strategy} during inference, which constructs the original noise for each frame from a shared noise.
This strategy significantly improves the quality and stability of the generated videos with negligible computational costs. 

We evaluate our \methodname~on several motion cases.
With a simple tuning using 8 $\sim$16 videos on a single GPU, the proposed \methodname~can generate videos with the common motion pattern of the video set and generalize well to unseen styles and objects. (See Figure~\ref{fig:results}).
Our key contributions can be summarized as follows:

\begin{itemize}
    \item We present a new setting of the few-shot tuning for the T2V generation task, aiming to strike a balance between generation freedom and training costs. 
    \item We propose \methodname, a baseline method for few-shot T2V called \methodname, which effectively learns the motion pattern in the given video set following a simple tuning.
    \item We introduce a first-frame conditioned pipeline that uses the first frame as a condition, effectively decoupling the motion and content, which simplifies the T2V task significantly. 
    \item We introduce the temporal layers and inference tricks, offering key insights into few-shot-based video generation. 
\end{itemize}
\section{Related Work}

\subsection{Text-to-Image Diffusion Models}
Recently, diffusion models~\cite{ho2020denoising, song2020denoising, song2020score} beat GANs~\cite{goodfellow2020generative, zhang2017stackgan, esser2021taming}, VAEs~\cite{kingma2013auto, sohn2015learning, van2017neural}, and flow-based~\cite{chen2019residual, grcic2021densely} approaches and have been in the limelight for text-to-image generation because of their stable training and outstanding performance.
For example, GLIDE~\cite{nichol2021glide} uses textural prompts as conditions and adopts classifier-free guidance~\cite{ho2022classifier} to improve image quality.
DALLE-2~\cite{ramesh2022hierarchical} introduces the pre-trained CLIP~\cite{radford2021learning} model to align the features of images and text.
Imagen~\cite{saharia2022photorealistic} injects the features from a large language model to diffusion models for better prompts understanding and proposes a cascaded pipeline to generate high-resolution images from coarse to fine.
To ease the computational burden of the iterative denoising process, Rombach \etal propose LDM~\cite{rombach2022high} that uses an autoencoder~\cite{kingma2013auto, esser2021taming} to reduce the redundancy of images. 
LDM compresses an image into low-dimension latent space by a pre-trained autoencoder first, then learns to denoise noisy latent data.
With the success achieved by LDM, many variants~\cite{peebles2023scalable, zhang2023adding} are proposed to improve the performance further.
More recently, the SD-XL~\cite{podell2023sdxl} is presented, which can generate extremely photo-realistic images with high-definition details.
In our work, SD-XL is utilized to generate the first frame, and SD-v1.4 is modified for subsequent frame prediction. 

\subsection{Text-to-Video Diffusion models}
The thriving of diffusion-based models in the text-to-image field demonstrates its potential in text-to-video generation.
The mainstream works can be divided into two categories: open-domain T2V generation and template-based methods.

\noindent\textbf{Open-domain T2V generation.}
During the early stage,  ImagenVideo~\cite{ho2022imagen} and Make-A-Video~\cite{singer2022make} learn T2V on the pixel level.
However, the video length and resolution are significantly limited due to the high computation in the pixel space.
MagicVideo~\cite{zhou2022magicvideo} is then proposed, which trains a new autoencoder on video data.
As the appearance of LDMs~\cite{rombach2022high} to the T2I field, MagicVideo boosts the computational effectiveness for T2V generation.
Blattmann \etal~\cite{blattmann2023align} present an LDMs-based T2V diffusion model, which adds extra 3D convolutional layers on frozen pre-trained layers.
VideoComposer~\cite{wang2023videocomposer} adds diverse conditions, \eg sketch and motion vectors, to the T2V model by a novel encoder.
AnimateDiff~\cite{guo2023animatediff} trains a set of motion layers capable of being applied to customized T2I models~\cite{hu2021lora, ruiz2023dreambooth}, enabling them to produce videos in a consistent style.
The above methods achieve remarkable performance for T2V generation.
However, the necessity of training these models on large-scale data like WebVid-10M~\cite{bain2021frozen} and HD-VILA-100M~\cite{xue2022advancing} poses a significant barrier for most researchers.
In addition, some zero-shot methods~\cite{khachatryan2023text2video,hong2023large,huang2023free} have been proposed, yet they often suffer from suboptimal frame consistency.

\noindent\textbf{Template-based methods.}
Template-based T2V generation aims to facilitate video-to-video translation with the guidance of user prompts, which is also known as video editing.
Dreamix~\cite{molad2023dreamix} and GEN-1~\cite{esser2023structure} are two pioneer works in template-based methods, while their training costs are comparable to open-domain T2V methods.
Then, Tune-A-Video~\cite{wu2023tune} proposes a new one-shot setting that uses a T2I model to overfit the origin video, which can be implemented on consumer-grade GPUs.
FateZero~\cite{qi2023fatezero} proposes a training-free method by injecting the cross-attention map of the source video and modifying attention layers.
Rerender-A-Video~\cite{yang2023rerender} and TokenFlow~\cite{geyer2023tokenflow} further improve the consistency of videos with the integration of priors and conditional guidance.
Different from the objectives of template-based methods, our few-shot T2V setting aims to achieve a higher degree of freedom in video generation rather than precisely aligning with the motion pattern of a template video.

\begin{figure*}[t]
    \centering
    \includegraphics[width=\textwidth]{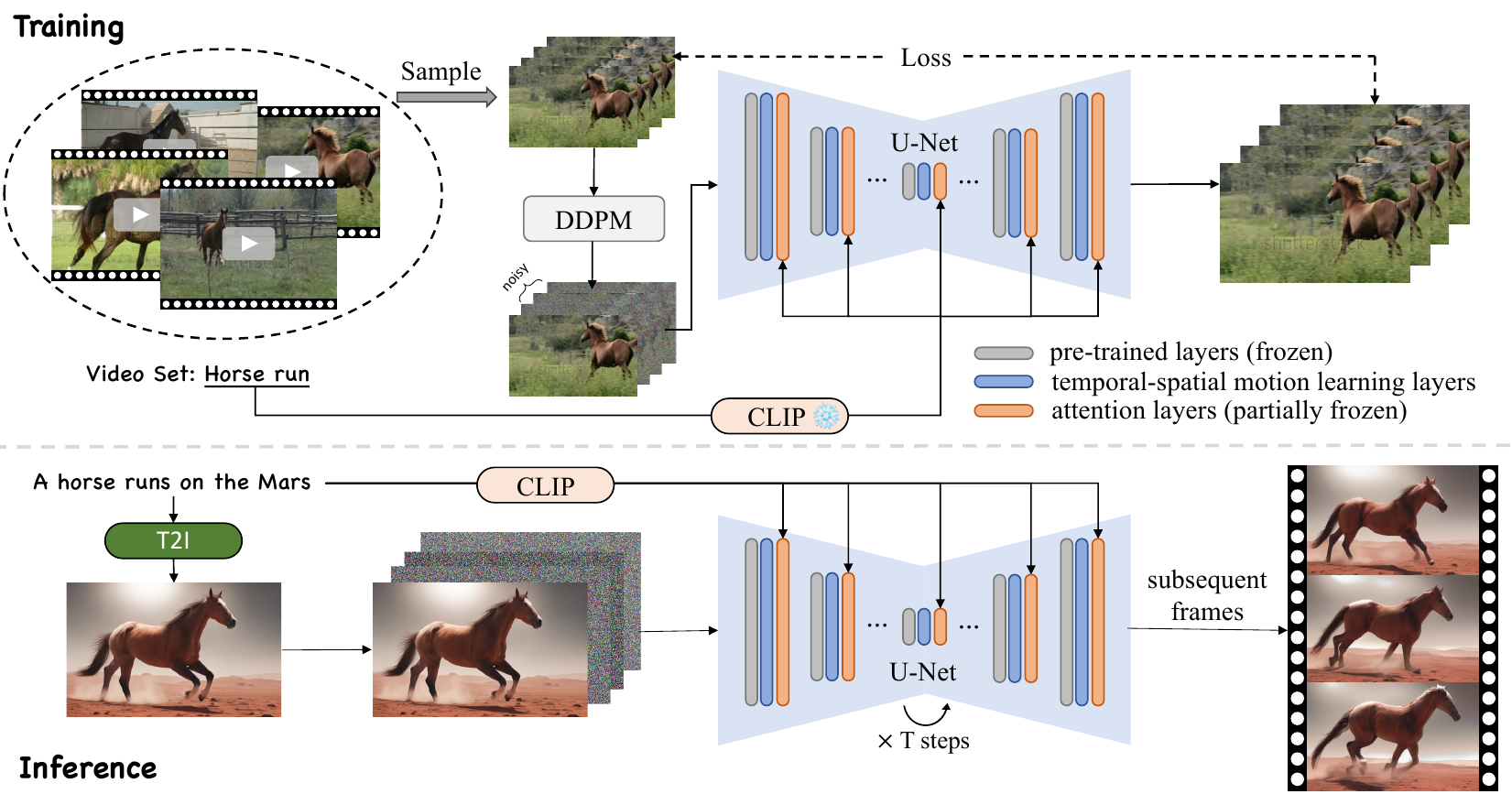} 
    \vspace{-0.6cm}
    \caption{\textbf{Framework of \methodname.} \methodname~learns a motion pattern from a small video set, enabling the generation of videos imbued with the learned motion patterns. This approach strikes a balance between training resources and generation freedom in video generation. We transfer text-to-video generation to the first-frame generation and subsequent-frame prediction, \ie, decoupling a video's contents and motions. During training, we add noise and compute loss functions for all frames except the first frame. Moreover, only the parameters of newly added layers and the query linear layers of self-attention blocks are tuned. During inference, we use a T2I model to generate the first frame. The tuned model only works on denoising the latent features of subsequent frames with the guidance of user prompts.}
    \label{fig:lamp}
    \vspace{-0.5cm}
\end{figure*}

\section{Method}
In this section, Sec.~\ref{sec:preliminaries} and Sec.~\ref{sec:setting} first introduce the preliminary knowledge and the new few-shot setting.
Next, Sec.~\ref{sec:first-frame-condition} details our proposed first-frame-conditioned pipeline.
Sec.~\ref{sec:architecture} is followed to describe how we modify a T2I diffusion model to T2V generation.
Finally, Sec.~\ref{sec:inference} introduces our shared-noise sampling strategy and some techniques that can improve performance during inference time.

\subsection{Preliminaries}
\label{sec:preliminaries}
In this section, we introduce the preliminary knowledge of the diffusion-based model.
Given  data $x_0 \in X$, a Markov chain can be defined as:
\begin{equation}
    q(x_t|x_{t-1}) = \mathcal{N}(x_t; \sqrt{1-\beta_t}x_{t-1}, \beta_tI),
\end{equation}
where $t=1,...,T$, $T$ is the total number of steps.
$\beta_t$ is a coefficient that controls the noise strength in step $t$.
The iterative noise adding can be simplified as:
\begin{equation}
    x_t=\sqrt{\bar{\alpha}_t}x_0 + \sqrt{1 - \bar{\alpha}_t}\epsilon, \quad\epsilon \sim \mathcal{N}(0, I),
\end{equation}
where $\bar{\alpha}_t = \prod_{i=1}^t(1-\beta_t)$. Diffusion models learn the distribution of dataset $X$ by minimizing the training objective,
which can be written as:
\begin{equation}
    \mathop{\arg\min}\limits_\theta\mathbb{E}_{x_0, \epsilon \sim \mathcal{N}(0,I), t, c}[||\epsilon - \epsilon_\theta(x_t, t, c)||_2^2],
    \label{eq:training_objective}
\end{equation}
$\epsilon_\theta(\cdot)$ denotes the noise prediction function of diffusion models, $c$ is the conditions like textual prompts.
After training, diffusion models can generate data from noise by reversing the noise-adding process.

However, the computational burden becomes substantial when diffusion models are used to generate high-resolution images.
To address this challenge, Latent diffusion models (LDMs) for T2I generation have been proposed, adopting an auto-encoder to achieve all operators in the latent space.
They can acquire low-redundancy initial data by 
encoder and reconstruct generated results by decoder.
LDMs are also used in our method to generate high-resolution videos.

\subsection{Our Few-shot-based T2V Generation Setting}
\label{sec:setting}
Existing T2V approaches require large-scale data for training or rely on a template video to obtain low-degree-of-freedom generative capabilities.
In order to make video generation inexpensive and flexible, we propose a novel setting: few-shot T2V generation.
Supposing that there is a video set $\mathbf{V}=\{\mathcal{V}_i | i \in [1, n]\}$ contains $n$ videos and a prompt $\mathcal{P}_m$  to describe the common motion as training data.
The proposed new setting is to tune a T2I model on the given video set and the motion prompt. 
The tuned model can generate a new video $\mathcal{V}^{\prime}$ with a similar motion pattern to $\mathbf{V}$ from a prompt $\mathcal{P}$ that is related to the motion.
We hope to learn the common motion pattern from a small video set while ignoring the contents.
%
%
Meanwhile, the training cost is affordable because of the small data size.
Based on the proposed setting, we modify pre-trained T2I models and present a baseline framework for few-shot T2V generation.

\subsection{First-Frame-Conditioned Pipiline}
\label{sec:first-frame-condition}
Due to the limited data in the few-shot tunning process, there is a  risk of overfitting the content of the small dataset, potentially compromising the degree of generation freedom.
To direct our model's focus toward motion, we propose the first-frame-conditioned pipeline to decouple motions and contents. The proposed pipeline is illustrated in Figure~\ref{fig:lamp}.
Based on our observation, the first frame contains the majority of the contents of a short video.
It is natural to use the first frame as a condition, enabling the model to pay more attention to motions.
Therefore, the T2V generation task is translated to first-frame T2I generation and subsequent-frame prediction.
There are previous works~\cite{esser2023structure, wang2023videocomposer} that have also used the first frame as a condition.
They concat it to the input noise or add a specific encoder to inject the features into networks.
However, applying these methods in the few-shot setting is challenging, as the limited data makes it nearly impossible to facilitate model training through substantial modifications to T2I models.
In contrast, the proposed first-frame-conditioned pipeline can achieve comparable effects with slight parameter changes, as detailed in Sec.~\ref{sec:architecture}.

Specifically, let $\mathcal{V} = \{f^i|i=1,...,l\}$ be a video contains $l$ frames and encode them into latent space: $\mathcal{Z} = \{z^i|i=1,...,l\}$.
When training the model, we preserve the original signal of $z^1$ and add noise to $\{z^2,...,z^l\}$.
The loss functions can be written as:
\begin{equation}
    \mathcal{L} = \mathbb{E}_{\mathcal{Z}, \epsilon \sim \mathcal{N}(0, I), t, c}[||\epsilon^{2:l}-\epsilon^{2:l}_\theta(\mathcal{Z}_t, t, c)||_2^2],
\end{equation}
where $\epsilon^{2:l}$ is the added noise from $2$nd to $l$-th frame, respectively.
Other notations are consistent with Eq.~\eqref{eq:training_objective}.
After training, the model gains the capability to generate a video with the motion pattern of the video set according to the first frame.
During inference, the powerful SD-XL~\cite{podell2023sdxl} is employed to provide the first frame $\hat{f}^1$, which is decoded to $\hat{z}^1$
Then, a sequence, $[\hat{z}^1, \epsilon^2,...\epsilon^l]$, where $\epsilon$ is a random noise, is fed to the model for the whole video generation.
At each step, we preserve the latent of the first frame and denoise the subsequent frames.

The proposed pipeline effectively avoids learning the contents of the video set so that it can train a model on limited data.
Another advantage lies in the quality of content generated by SD-XL, providing a good reference for video generation. This approach enables us to leverage the advantages of well-established T2I techniques.
%
%
The first-frame-conditioned pipeline significantly benefits both prompt alignment performance and generation diversity.
Moreover, this pipeline is also appealing in its flexibility in applications \eg real-world image animation and video editing, as detailed in Sec.~\ref{sec:more-application}.

However, the original T2I models treat frames as independent samples.
Thus, the features of the first frame cannot be used to establish temporal relationships between frames and generate videos.
The following section introduces how we enable the model to work at the temporal level.

\subsection{Adapt T2I Models to Video}
\label{sec:architecture}
\begin{figure}[t]
    \centering
    \includegraphics[width=\linewidth]{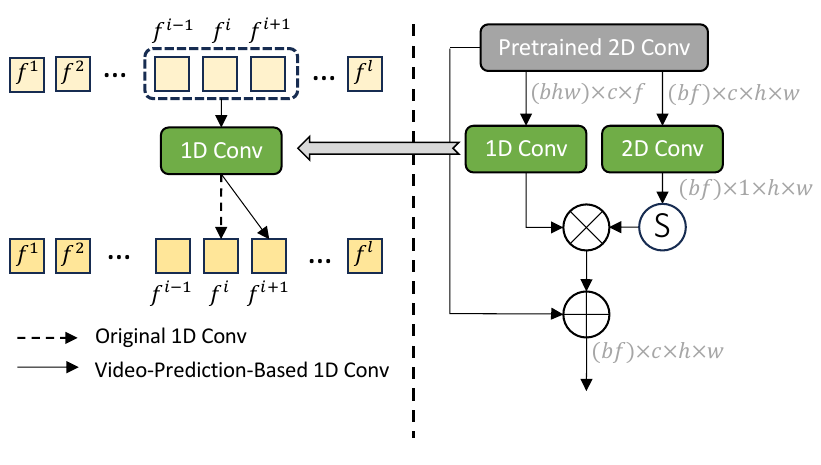}
    \put(-226, 5){\small{(a)~The proposed 1D conv}}
    \put(-113, 10){\small{(b)~Temporal-spatial motion}}
    \put(-83, -2){\small{learning layer}}
    \caption{The details of the proposed temporal-spatial motion layers.
    (b) illustrates that 1D convolutions are added on pre-trained layers to capture information along the temporal dimension.
    2D convolution layers with an output channel number of $1$ control the spatial level's motion strength. The 1D convolutional layers utilize the former two frames to generate the current frame, as shown in (a). $f^i$ denotes the $i$-th frame.}
    \vspace{-0.5cm}
    \label{fig:convlayer}
\end{figure}

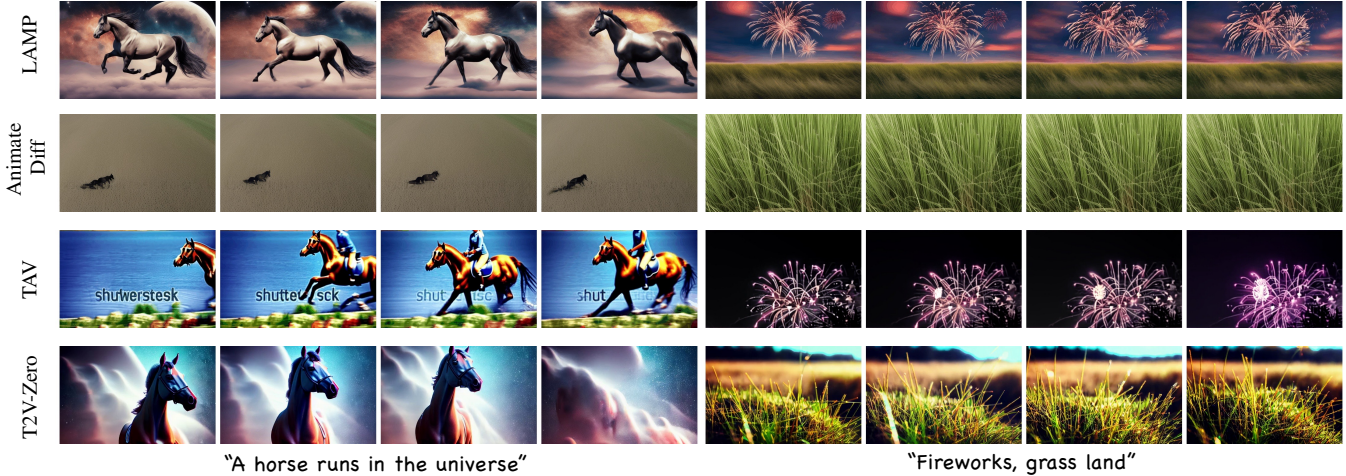
\begin{figure*}[t]
    \flushright
    \vspace{-0.1cm}
    \includegraphics[width=0.985\textwidth]{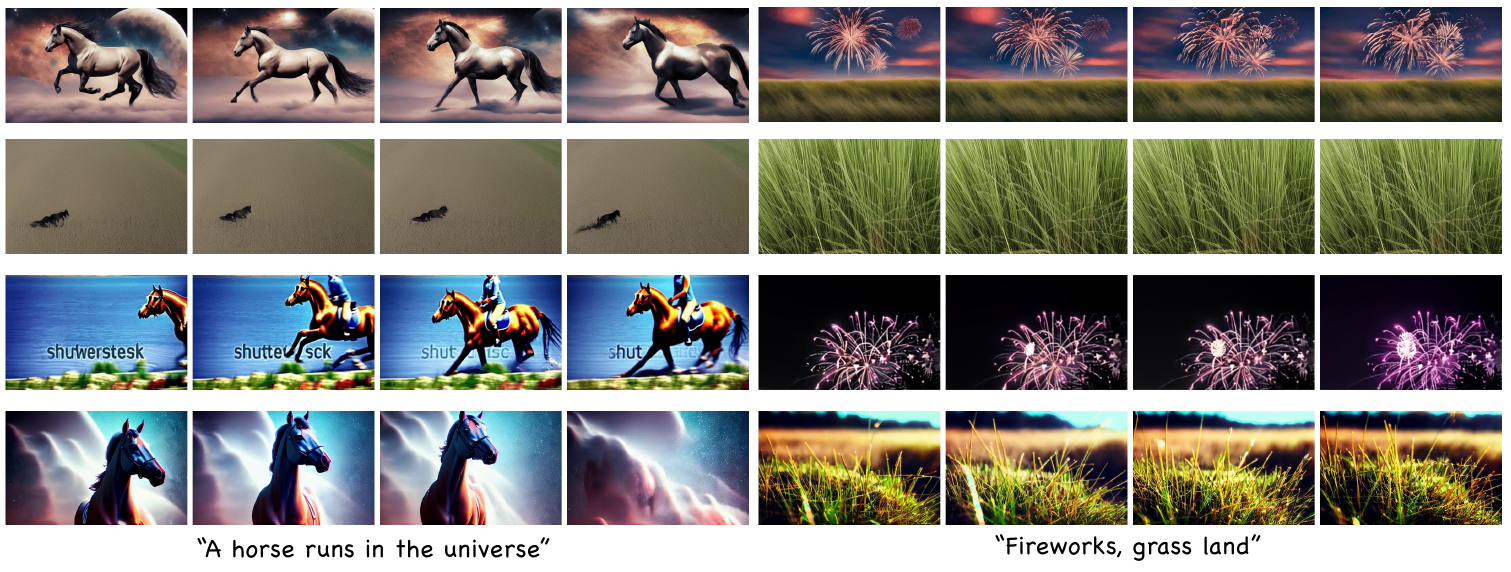}
    \put(-502, 155){\footnotesize{\rotatebox{90}{\methodname}}}
    \put(-508, 109){\footnotesize{\rotatebox{90}{Animate}}}
    \put(-499, 117){\footnotesize{\rotatebox{90}{Diff}}}
    \put(-502, 70){\footnotesize{\rotatebox{90}{TAV}}}
    \put(-502, 17){\footnotesize{\rotatebox{90}{T2V-Zero}}}
    \vspace{-0.35cm}
    \caption{Qualitative comparison between the proposed \methodname~and three baselines. \textbf{Zoom in for the best view.}}
    \label{fig:comparison}
    \vspace{-0.38cm}
\end{figure*}

\noindent\textbf{Temporal-spatial motion learning layers.}
To empower the T2I model to extract temporal features, we inflate the pre-trained 2D convolutional layers into the proposed temporal-spatial motion learning layers.
As illustrated in Figure~\ref{fig:convlayer}(b), the proposed layer consists of two branches.
Suppose the latent features of the input video are represented as as a 5D tensor with a shape of $b\times c \times f \times h \times w$.
In the temporal branch, the tensor is reshaped into $bhw \times c \times f$ and fed to a 1D convolutional layer.
However, since the 1D convolution kernel can only work on a spatial coordinate at a time, it fails to take the essential spatial features into account.
Consequently, a 2D convolution with an output channel of $1$ along with a Sigmoid function is added as compensation for spatial features.
The input features are reshaped into $bf \times c \times h \times w$ in the spatial branch.

Considering that our first-frame-conditioned pipeline needs to predict subsequent frames based on the given first frame,
which is similar to video prediction~\cite{lotter2016deep, hu2023dynamic}, we design our 1D convolutional layers in a video prediction manner, as shown in Figure~\ref{fig:convlayer}(a).
When the kernel slides through the features of frames $\{f^{i-1}, f^i, f^{i+1}\}$, our \textbf{video-prediction-based 1D convolution} produces the features of $f^{i+1}$ instead of $f^i$ as in the original version.
Thus, we can utilize the former two frames to predict the subsequent frame, \ie effectively achieving video prediction in the base operators.
Moreover, to avoid our newly added layers  polluting the generation capability of the original T2I model,
all parameters are zero-initialized, as done in ControlNet~\cite{zhang2023adding}.

\noindent\textbf{Attention layers.}
We also modify the attention layers to ensure consistency.
For self-attention layers, all key and value features are obtained from the first frame, which can be written as:
\begin{equation}
    {\rm Attention}(Q^i, K^1, V^1) = {\rm Softmax}(\frac{Q^i(K^1)^T}{\sqrt{d}})V^1,
\end{equation}
the superscript $i \in \{1, ..., l\}$ indicates the features are from the $i$-th frame.
Combined with the proposed pipeline, the reformulated self-attention layers facilitate subsequent frames to refer back to the conditions established by the first frame.
Besides,  it has been demonstrated that such a modification can effectively preserve the main object even without tuning \cite{khachatryan2023text2video}.
Moreover, following the modification in~\cite{wu2023tune}, we have incorporated  temporal attention layers, which are self-attention
layers working on the temporal dimension.

\subsection{Shared-noise Sampling During Inference}
During inference, we propose a simple yet effective shared-noise sampling strategy further to improve the quality of the generated videos quality.
Specifically, we first sample a shared noise $\epsilon^s \sim \mathcal{N}(0,I)$.
Then, a noise sequence $[\epsilon^2,...,\epsilon^l]$ with the same distribution as the base noise is sampled.
In our sampling strategy, the original noise $\epsilon^i$ for the $i$-th frame generation is updated as:
\begin{equation}
    \epsilon^i = \alpha \epsilon^s + (1 - \alpha) \epsilon^i,
\end{equation}
where $\alpha$ is a coefficient to control the sharing degree.
We empirically set $\alpha=0.2$ in our experiments.
This approach ensures a consistent noise level across each frame, ultimately manifesting as consistency in the generated videos.
Intuitively, this approach is in accordance with the prior knowledge that every frame of a video has certain similarities.
Mathematically, the reduced noise variance can contract the dynamic range of the latent space, contributing to a more stable generation process.
Besides, the AdaIN~\cite{huang2017arbitrary} technique on latent space and histogram matching at the pixel level are used for post-processing.
The efficacy of our free-lunch inference strategies will be demonstrated in Sec.~\ref{sec:ablation}.

\label{sec:inference}
\section{Experiments}

\subsection{Implementations}
In our experiments, we generate videos with resolutions of $320 \times 512$ and $16$ frames.
%
We use SD-XL~\cite{podell2023sdxl} for the less computationally intensive first frame generation and the relatively more lightweight SD-v1.4~\cite{rombach2022high} for the more computationally demanding prediction of subsequent frames, thereby balancing the inference cost of the two stages.
For training, we use a set of self-collected videos ranging from $8 \sim 16$,  randomly sampling a $16$-frame clip during each iteration
%
All frames are resized to a resolution of $320 \times 512$.
Only the parameters of new-added layers and the query linear layer in self-attention blocks are tuned, and the learning rate is set to $3.0 \times 10^{-5}$.
All experiments are implemented on a single A100 GPU and only need $\sim15$ GB vRAM for training and $\sim6$ GB vRAM for inference.

\subsection{Comparisons}
We train our~\methodname~for $8$ motions, including helicopter (rigid motion), waterfall (fluid motion), rain \& firework (particle motion), horse running (animal motion), birds flying (multi-body motion), turn to smile (human emotion) and play the guitar (human motion). We design $6$ prompts for each motion to build an evaluation set containing $48$ videos.
Three publicly available methods, which are large-scale pre-trained AnimateDiff~\cite{guo2023animatediff}, one-shot-based video editing method Tune-A-Video~\cite{wu2023tune}, and zero-shot-based Text2Video-Zero~\cite{khachatryan2023text2video}, are selected as our comparison baselines. 
%
We consider representative work under a variety of mainstream settings, thus effectively reflecting the advantages of our few-shot learning setting.
Notably, for each motion pattern, we randomly select a video from the corresponding video set as the template to train Tune-A-Video~\cite{wu2023tune}.
The comparisons are constructed in the view of objective and subjective.

\begin{table}[t]
    \centering
    \footnotesize
    \caption{\textbf{Quantitative comparisons with the evaluated text-to-video methods.}}
    \begin{tabular}{c|ccc}
    \toprule
        Method & Alignment~$\uparrow$ & Consistency~$\uparrow$ & Diversity~$\downarrow$ \\ \midrule
        Tune-A-Video~\cite{wu2023tune} & 27.2227 & 94.8742 & 84.7186\\
        T2V-Zero~\cite{khachatryan2023text2video} & 26.9424 & 91.4713 & 73.0136\\
        AnimateDiff~\cite{guo2023animatediff} & 28.8779 & 97.8131 & 73.4723\\ \midrule
        LAMP~(Ours) & \textbf{31.3547} & \textbf{98.3085} & \textbf{71.6535}\\ \bottomrule
    \end{tabular}
    \label{tab:quantitative}
\end{table}

\begin{figure}[t]
    \centering
    \includegraphics[width=\linewidth]{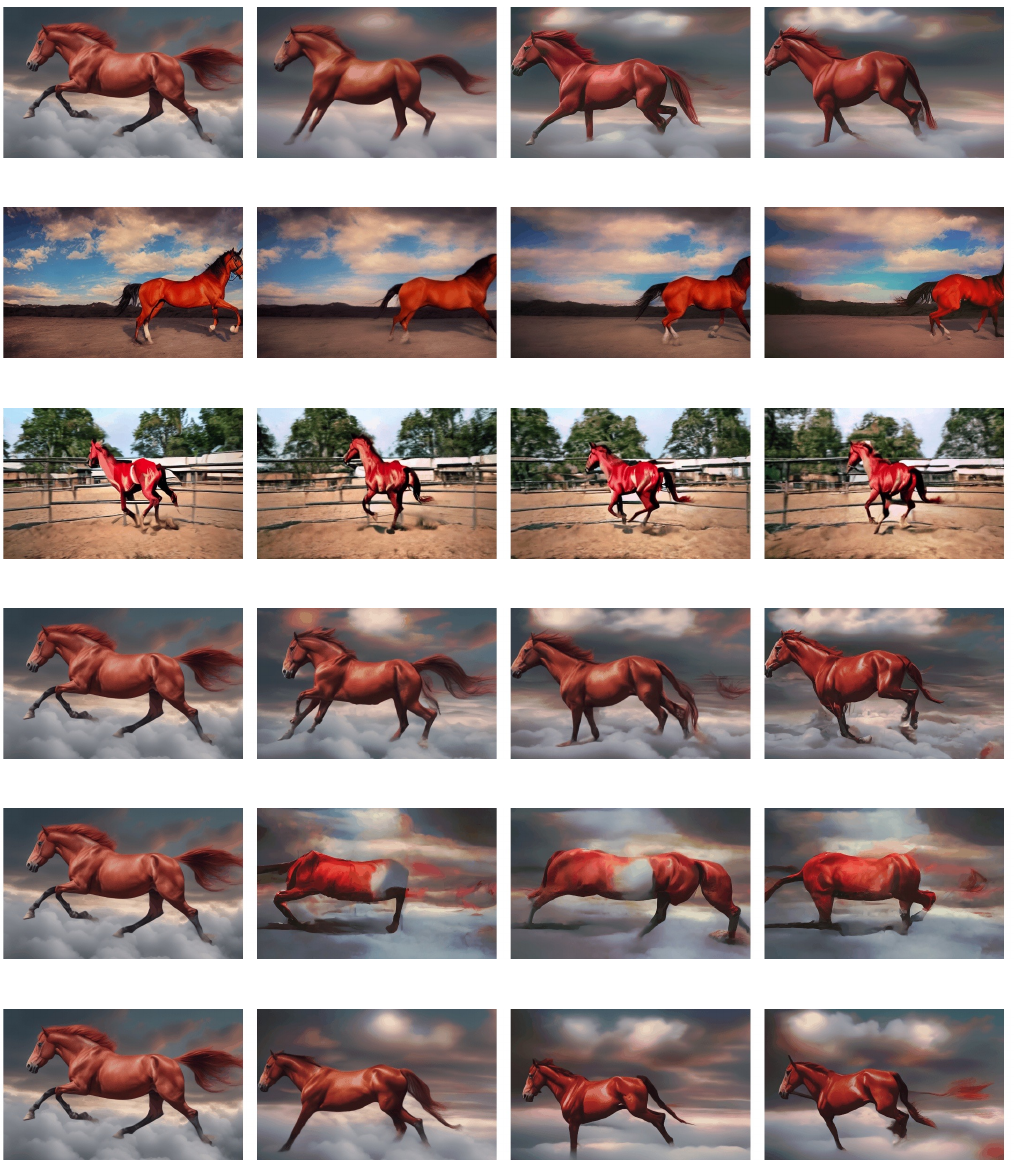}
    \put(-162, 227.5){\small{(a)~\methodname~(full model)}}
    \put(-174, 180.5){\small{(b)~replace SD-XL to SD-v1.4}}
    \put(-190, 135){\small{(c)~w/o first-frame-conditioned pipeline}}
    \put(-175, 87.5){\small{(d)~w/o shared-noise sampling}}
    \put(-207, 40.5){\small{(e)~w/o temporal-spatial motion learning layer}}
    \put(-195, -7.5){\small{(f)~w/o video-prediction-based 1D Conv}}
    \vspace{-0.15cm}
    \caption{\textbf{Ablation results.} The given prompt is `A red horse runs in the sky'.}
    \label{fig:ablation}
    \vspace{-0.45cm}
\end{figure}

\begin{figure*}
    \includegraphics[width=\textwidth]{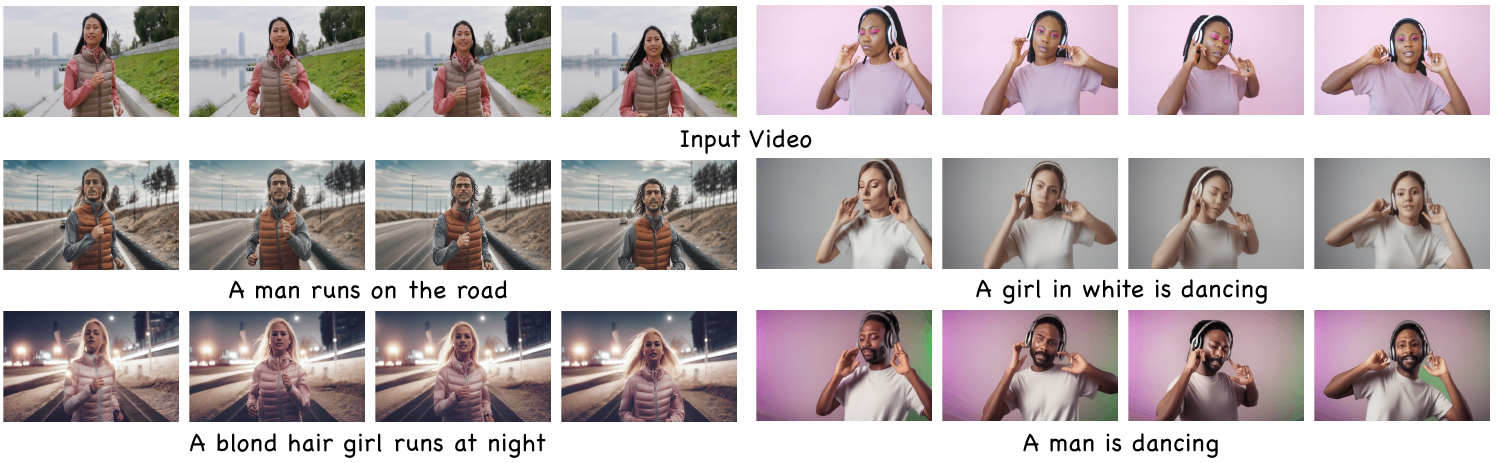}
    \vspace{-0.6cm}
    \caption{Visual results of our video editing application.
    \textbf{Zoom in for the best view.}}
    \label{fig:video-editing}
    \vspace{-0.35cm}
\end{figure*}

\begin{figure}
    \centering
    \includegraphics[width=\linewidth]{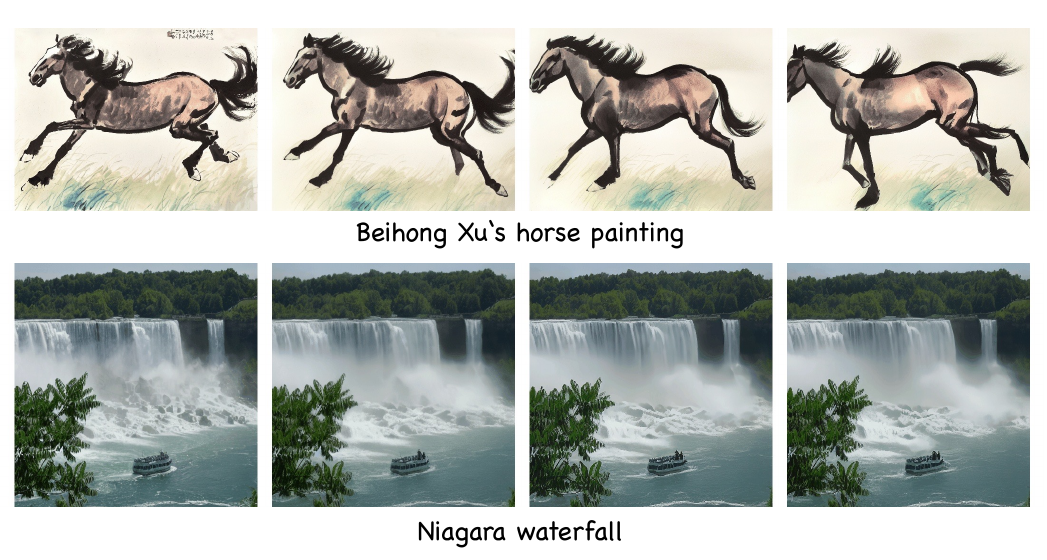}
    \vspace{-0.7cm}
    \caption{Visual results of \methodname~animates the real-world images.}
    \vspace{-0.55cm}
    \label{fig:real-image-animation}
\end{figure}

\noindent\textbf{Quantitative results.}
We evaluate our \methodname~against baselines in terms of textural alignment, frame consistency, and generation diversity.
The objective metrics and user study are used for a comprehensive evaluation.

\textit{Objective metrics.}
To measure the textual alignment of a video, we average each frame's CLIP score~\cite{radford2021learning}.
Following~\cite{wu2023tune}, we also represent the frame consistency by the mean cosine similarity of  CLIP image embedding across all frame pairs.
Since generation freedom is one of our core goals, we also include generation diversity in quantitative evaluation.
We use the average CLIP image embedding of all frames to represent a video, subsequently computing and averaging the cosine distance across all video pairs. A lower score denotes lower similarity, \ie, better diversity.
Table~\ref{tab:quantitative} presents the quantitative results of~\methodname~and baselines.
Across all three evaluation criteria,  our method achieves state-of-the-art performance against the other baselines.

\textit{User study.}
We further conduct a user study to evaluate our approach and three baselines subjectively.
We randomly select 24 cases from our evaluation set. In each case, we ask the participant ``Which video do you think has better visual quality and better matches the scene and motion of the prompt `...'?" 
The user study garnered a total of 70 responses from a diverse group of participants, including both experts in the field and individuals with no specific background knowledge. 
Statistically, $46.84\%$ of respondents favor our method, with AnimateDiff~\cite{guo2023animatediff} achieving $19.11\%$ and  Tune-A-Video~\cite{wu2023tune} achieving
$22.15\%$.
However, it is worth noting that Tune-A-Video polarizes choices in different situations. 
When there are similarities between the layout of the given video template and the scene described by the prompt, combined with its own good frame consistency, it can be approved by most volunteers.
Conversely, the textural alignment of its generated video is poor, \eg "Fireworks, grass land" shown in Figure~\ref{fig:comparison}.
Besides, $11.90\%$ of the participants select Text2Video-Zero~\cite{khachatryan2023text2video} as their preference.
As a result, our~\methodname~obtains the highest approval rate among the participants.

\noindent\textbf{Qualitative results.}
We present several visual examples of our method and three baselines in Figure~\ref{fig:comparison}. 
AnimateDiff~\cite{guo2023animatediff} learns motion layers on large-scale data and inserts them into personalized T2I models to generate videos with specific styles and better visual quality. 
However, this approach cannot be combined with the better-performing but heterogeneous T2I model, resulting in a limitation in textural alignment capabilities even though the consistency and diversity are satisfying. 
This limitation is apparent in cases `A horse runs in the universe' and `Fireworks, grass land'. 
Tune-A-Video (TAV) can only generate videos with the same motion, with the prompts sometimes unable to effectively control the generated videos due to overfitting on the given video. 
While T2V-Zero produces visually pleasing frames, it falls short in generating videos with meaningful motion patterns.
In contrast, our \methodname~achieves good consistency and generates videos with proper motion patterns, benefiting from the proposed motion learning layers.
Besides, using the advantage of our first-frame-conditioned pipeline, the proposed method achieves visual quality on par with state-of-the-art T2I models, even with the modifications based on SD-v1.4.
Figure~\ref{fig:results} and supplementary materials provide more visual results. Our method understands the learning motions well and can generalize to diverse, even unseen, scenes and styles.

\subsection{Ablation Study}
\label{sec:ablation}
We conduct ablation experiments to demonstrate the effectiveness of each proposed component.
The visual results are shown in Figure~\ref{fig:ablation}. 
%
As we can see in Figure~\ref{fig:ablation}(b), using SD-v1.4 to generate the first frame will decrease the performance compared to the full model. 
Upon comparing Figure~\ref{fig:ablation}(c) with the video generated by the full model, the model without the first-frame-conditioned pipeline produces low-quality results.
Notably, the presence of unrelated objects, such as fences and dirt, in the video indicates an overfitting of the content of the video set.
In addition, the model w/o shared-noise sampling can generate relatively consistent frames but lacks smoothness in the result.
When the temporal-spatial motion learning layers are removed, the model cannot effectively capture the complex motion pattern, leading to failed results.
Finally, when we turn video-prediction-based 1D convolution into the original version of 1D convolution, the main object of the video becomes inconsistent.
The proposed layer can effectively preserve and propagate the features of the first frame to the subsequent frames, as depicted in the results.
These results verify the significant contributions of each key module to the final full model.
\vspace{-0.1cm}
\section{More Applications}
\label{sec:more-application}
In this section, we provide more applications of the proposed \methodname.
While primarily designed for text-to-video generation, our framework can also be used for real image animation and video editing.

\subsection{Real Image Animation}
Through the training of the proposed first-frame-conditioned pipeline, our \methodname~contains a network that predicts the subsequent frames based on the given first frame.
This enables the animation of real-world images generated by T2I models.
Thus, our method naturally gains the capability to animate real-world images based on the learned motion patterns if these images are placed in the first frame.
%
%
Figure~\ref{fig:real-image-animation} shows several representative cases in which the `horse run' model animates a famous horse painting created by Beihong Xu and the `waterfall' model makes the wonderful Niagara waterfall flow.
This application further demonstrates our generalization performance, even when dealing with complex real-world scenes.

\subsection{Video Editing}
In cases where the given training set contains only a single video clip, our method can only learn a specific motion rather than a motion pattern.
In this special case, our method effectively turns into a video editing algorithm.
The training process remains similar to that in the few-shot setting.
During inference, we adopt the ControlNet~\cite{zhang2023adding} based on SD-XL~\cite{podell2023sdxl} and condition it on canny edges to edit the first frame.
DDIM inversion~\cite{wu2023tune} is also used to provide a base motion, thereby ensuring better subsequent-frame prediction.
Similarly to video generation, our approach can also take full advantage of image-editing technologies when applied to video editing.
As visual examples shown in Figure~\ref{fig:video-editing}, our \methodname~generates photo-realistic videos while maintaining good frame consistency.

\section{Limitation and Future Works}
In our experiments, we observed that the occurrence of failure cases increased as our method attempted to learn complex motions.
More effective modules for motion learning are potential solutions to this issue. 
Besides, we found that the motion of the foreground object sometimes influences the background's stability.
We believe that learning the foreground and background movements independently might be an effective solution. 
We leave these improvements in our future work.

\section{Conclusion}
This paper proposes a novel setting, few-shot tuning for T2V generation, which learns a common motion pattern from a small video set to achieve a trade-off between training burden and generation freedom.
The proposed  \methodname~ serves as a baseline for this new setting.
In our method, we transfer the T2V task into T2I generation for the first frame and predict the subsequent frames.
This avoids overfitting the content of the dataset during few-shot tuning while leveraging the advantages of text-to-image techniques. 
Moreover, our novel design in network architecture and inference strategy further boosts the performance of T2V generation.
Extensive experiments demonstrate the effectiveness and generalization capability of our method.
We believe that the few-shot tuning setting offers superior trade-offs and will aid the broader T2V field in exploring the lower bounds on the data required for video diffusion training.
{
    \small
    \bibliographystyle{ieeenat_fullname}
    \bibliography{main}
}

\end{document}